\documentclass[11pt]{article}
\RequirePackage[colorlinks,citecolor=blue,urlcolor=blue,linkcolor=blue]{hyperref}
\usepackage{url}
\usepackage{booktabs}
\usepackage{svg}
\usepackage{tcolorbox}
\usepackage{subcaption}
\usepackage{url,amscd,paralist,bbm,wrapfig}
\usepackage{bm}
\usepackage{algorithm}
\usepackage{algorithmic} 
\usepackage{todonotes}
\usepackage{amsmath}
\usepackage{amsthm}
\usepackage{amsopn}
\usepackage{amsfonts}
\usepackage{cleveref}
\usepackage{natbib}

\title{What Makes a Good Example? \\ Modeling Exemplar Selection with Neural Network Representations}
 
\author{Fanxiao Wani Qiu\thanks{Department of Psychology, University of Southern California (email: \texttt{fanxiaoq@usc.edu})}\ $^\sharp$
  \and Oscar Leong\thanks{Department of Statistics and Data Science, University of California, Los Angeles (email: \texttt{oleong@stat.ucla.edu})}\ $^\sharp$
  \and Alexander LaTourrette\thanks{Department of Psychology, University of Southern California (email: \texttt{latourre@usc.edu})}}
\date{}

\begin{document}

\maketitle

\begin{abstract}

Teaching requires distilling a rich category distribution into a small set of informative exemplars. Although prior work shows that humans consider both representativeness and diversity when teaching, the computational principles underlying these tradeoffs remain unclear. We address this gap by modeling human exemplar selection using neural network feature representations and principled subset selection criteria. Novel visual categories were embedded along a one-dimensional morph continuum using pretrained vision models, and selection strategies varied in their emphasis on prototypicality, joint representativeness, and diversity. Adult participants selected one to three exemplars to teach a learner. Model–human comparisons revealed that strategies based on joint representativeness, or its combination with diversity, best captured human judgments, whereas purely prototypical or diversity-based strategies performed worse. Moreover, transformer-based representations consistently aligned more closely with human behavior than convolutional networks. These results highlight the potential utility of dataset distillation methods in machine learning as computational models for teaching.

\end{abstract}

\def\thefootnote{$\sharp$}\footnotetext{These authors contributed equally to this work.}

\noindent Teaching is a process central to cumulative culture. Infants are sensitive to teaching from infancy, treating information taught to them by adults as generalizable and culturally relevant \citep{csibra2011natural}. By preschool age, children themselves can engage in selective teaching, prioritizing transmitting optimally helpful information. This sensitivity to teaching, both as benefactors and beneficiaries, supports the accumulation of knowledge across generations and underlies humans' uniquely rich representational systems, including language, art, and science. An important part of teaching involves the selection of informative examples \citep{shafto2014rational}. For instance, when teaching a child about birds, one might weigh both representativeness (e.g., robins and sparrows) and diversity (e.g., penguins and ostriches) when deciding which exemplars best convey the category structure. 

Despite the importance of selecting good examples for successful teaching, there has been relatively little systematic work that examines \textit{how} humans select examples. Prior work suggests that adults do not simply select at random. Shafto and colleagues formalized teaching as a process of Bayesian inference, in which teachers choose examples to maximize a learner's posterior belief in the accurate hypothesis, rather than to reflect the underlying data distribution \citep{shafto2014rational}. Related work shows that adults prefer informative positive examples when teaching concepts \citep{avrahami1997teaching}, and that both children and adults favor diverse over similar examples when teaching others about a novel property in animals \citep{rhodes2010children}. These findings suggest that human teachers are sensitive to the informativeness of examples. However, the computational principles governing how humans resolve tradeoffs between representativeness and diversity in selecting examples of a category remain poorly understood. 

A parallel set of questions has been explored in machine learning under data compression and dataset distillation. Motivated by computational considerations and the observation that large datasets are often redundant, this body of work asks whether a small subset of curated examples can capture the essential structure of a larger dataset. Early works along these lines include core-set construction \citep{tsang2005core, har2007smaller}, facility location techniques \citep{mirchandani1990discrete}, and instance selection methods \citep{olvera2010review}, which seek to select a subset of training examples such that models trained on this subset achieve performance comparable to those trained on the full dataset. These techniques continue to be explored in the context of active learning \citep{wei2015submodularity, sener2018active, li2023bal}. Building upon these ideas, more recent work has explored the synthetic generation of small training sets that are explicitly optimized to encode the information contained in a larger dataset \citep{wang2018dataset}. This line of work, commonly referred to as dataset distillation \citep{lei2023comprehensive}, has been shown to be beneficial in reducing memorization \citep{sucholutsky2021secdd}, accelerating model evaluation and architecture search \citep{zhao2020dataset}, and addressing privacy concerns \citep{dong2022privacy}.


The variety of computational approaches to the task of selecting informative examples for learners raises new questions for human teaching strategies as well. Using criteria such as diversity, joint representativeness, or a combination of the two, work in dataset-distillation formalizes different possible strategies driving human pedagogical sample selection. For example, when selecting examples of the ``mammal" category, teachers might select diverse exemplars to highlight the extremes (e.g., bats and whales), prototypical exemplars to illustrate the central mass of the distribution (e.g., dogs and cats), representative exemplars to illustrate variation in non-extreme category members (e.g., otters and elephants), or some combination of these. Yet, existing accounts of human teaching have not systematically evaluated which objectives best characterize exemplar selection. 

\section{Our Contributions}

In this work, we aim to bridge this gap by leveraging tools from machine learning and dataset distillation to computationally model how humans select exemplars to teach others. We embed datasets using pretrained, neural network feature representations and formalize multiple candidate selection strategies that differ in how they balance concepts such as representativeness of the overall dataset, prototypicality, and diversity. These strategies define concrete metrics for choosing a subset of exemplars from a larger set. We then test adult human participants on the same selection task and evaluate which computational metrics best capture human behavior.

By grounding models of human exemplar selection in learned representations and principled subset selection objectives, our approach offers a new way to connect machine learning methodologies of dataset distillation with cognitive theories of teaching and information transmission. More broadly, this work suggests that advances in representation learning can help serve as a lens for understanding how humans structure information to support learning in others.

\section{Methods}

\subsection{Stimuli} 

We used visual stimuli created by \citep{havy2016naming}, including members of 3 novel categories (labeled as "daxes", "veps," and "bems"). In each category, members lie on a one-dimensional spectrum between two distinct creatures, continuously morphing various attributes such as color, feature, shape, and size. Each member is assigned a \textit{scale} (0 -- 100) that quantifies where on the spectrum it lies. These categories have previously been shown to be learnable by 2-year-old children \citep{latourrette2022sparse}.

\begin{figure*}[t]
  \centering
  \begin{subfigure}[t]{0.32\textwidth}
    \centering
    \includegraphics[width=\linewidth]{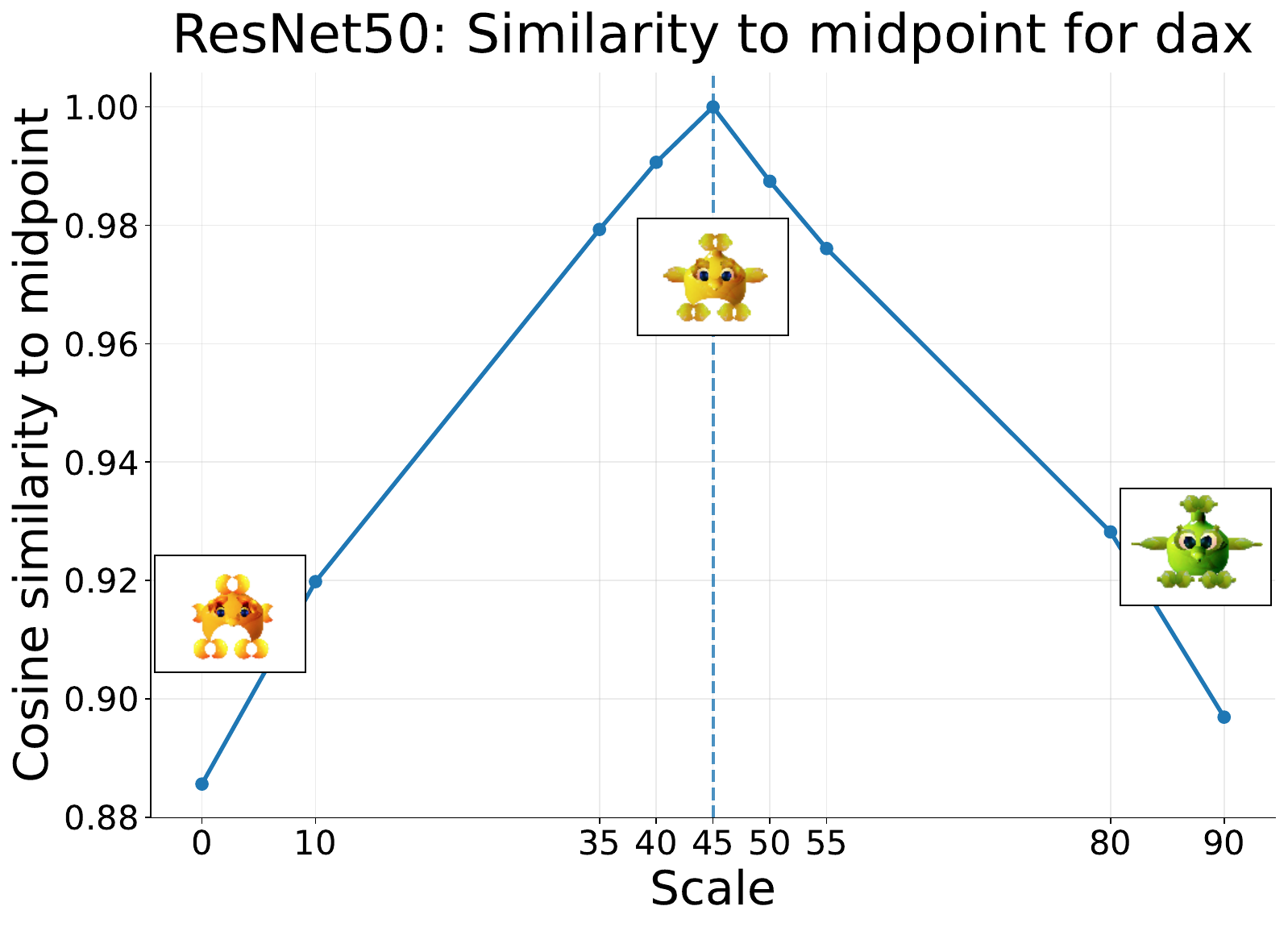}
  \end{subfigure}\hfill
  \begin{subfigure}[t]{0.32\textwidth}
    \centering
    \includegraphics[width=\linewidth]{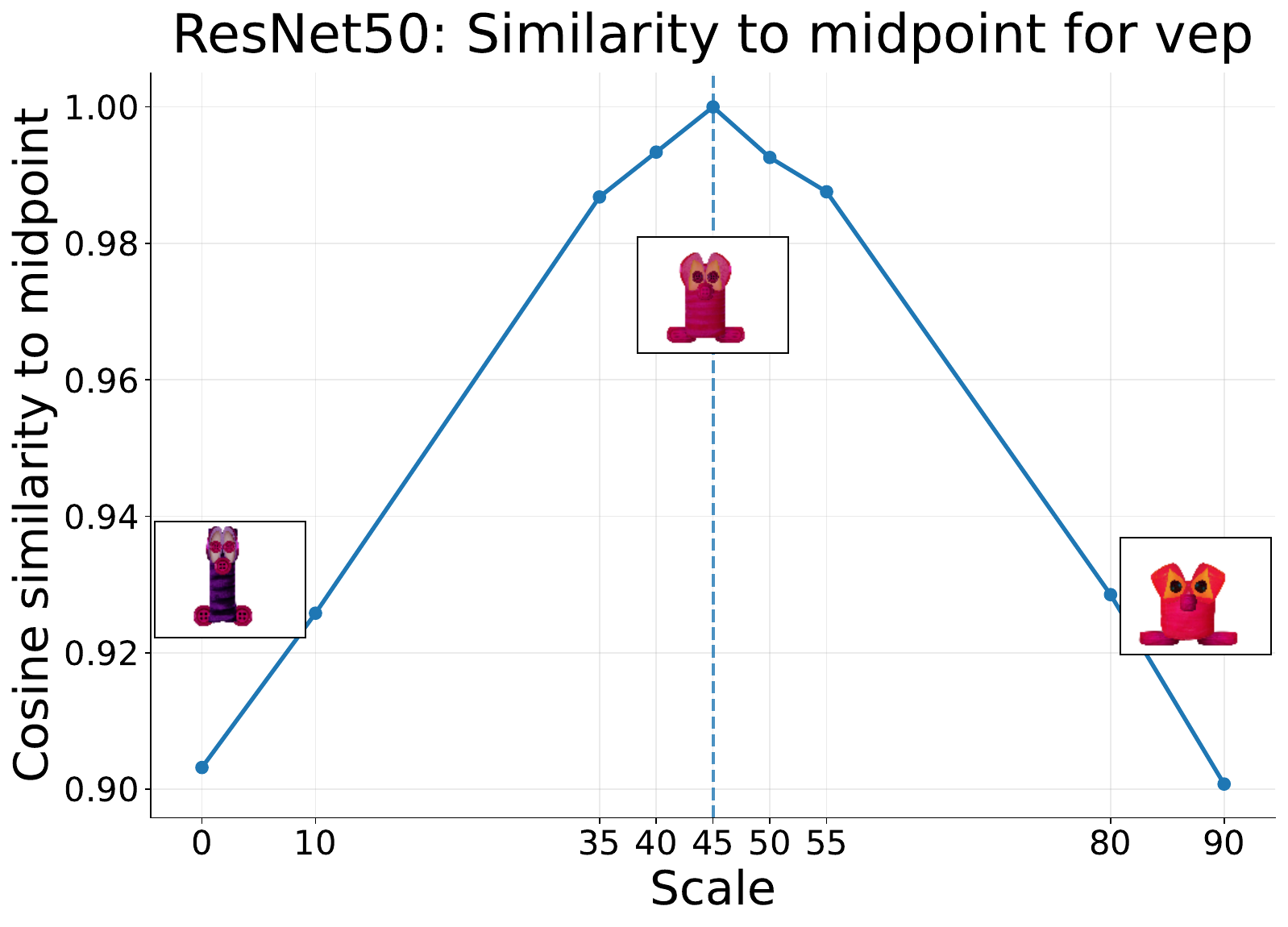}
  \end{subfigure}\hfill
  \begin{subfigure}[t]{0.32\textwidth}
    \centering
    \includegraphics[width=\linewidth]{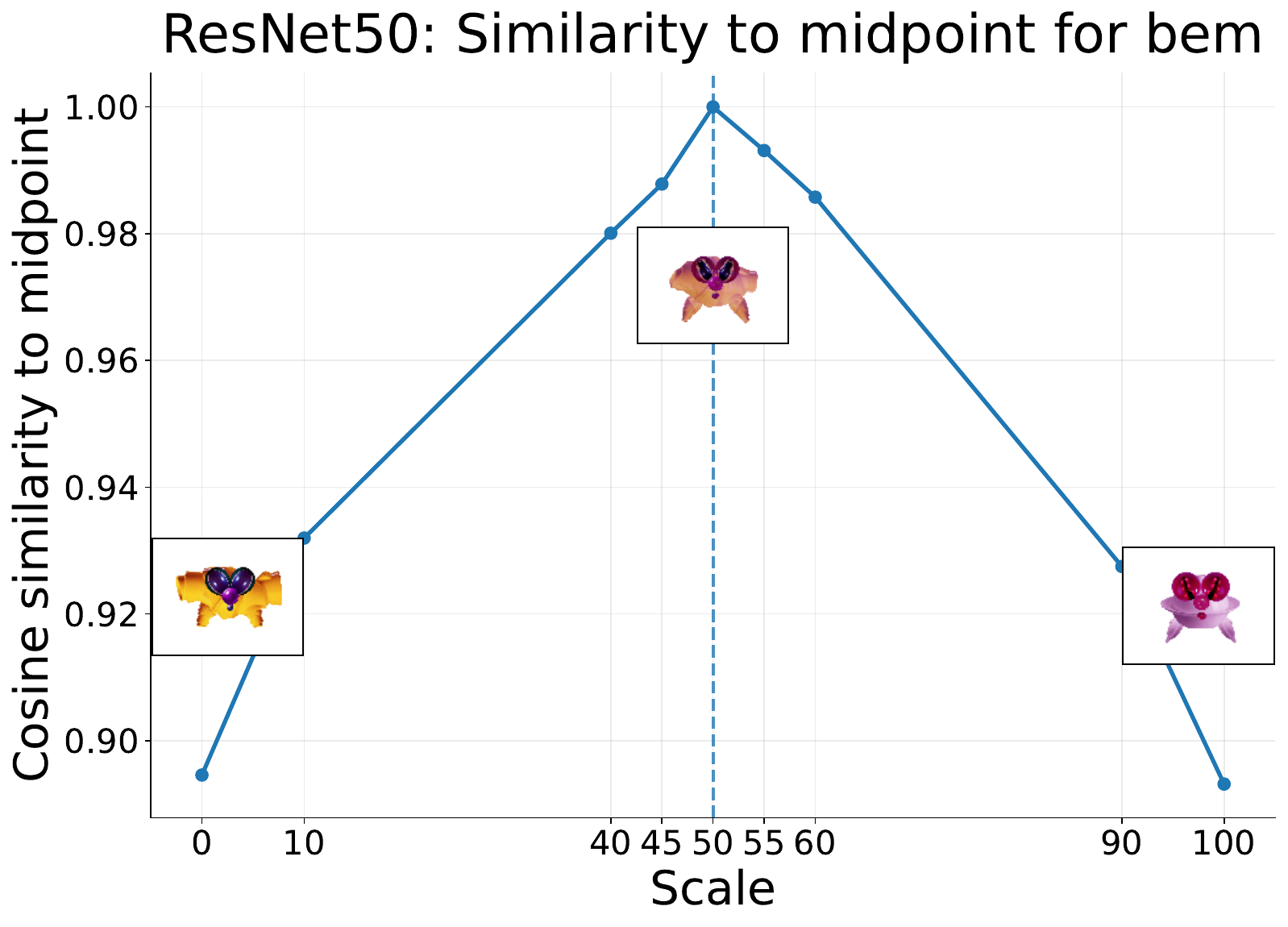}
  \end{subfigure}

  \caption{For each stimulus category (dax, vep, bem), we compute the cosine similarity between the ResNet feature representation of each member and the representation of the midpoint stimulus (dashed vertical line). We also visualize the left and right endpoints along with the midpoint.} 
  \label{fig:similarity-all-categories}
\end{figure*}

We then selected a range of exemplars from each category, ensuring that each category varied smoothly across exemplars and categories were roughly aligned in their within-category similarity. As shown in Figure~\ref{fig:similarity-all-categories}, the cosine similarity between each exemplar and the category midpoint peaks at the center of the scale and decreases smoothly and symmetrically as exemplars become more extreme. Thus, the midpoint is genuinely prototypical with respect to the chosen range and deviations toward either endpoint are balanced (see Figure \ref{fig:similarity-all-categories} for these ranges). 

\subsection{Neural Network-based Distillation}

To characterize how different neural network architectures prioritize representativeness, prototypicality, and diversity, we select exemplar subsets based on alignment in learned feature representations by a pretrained neural network. In particular, given $N$ images $x_1,\dots,x_N$ from a novel object category, we extract feature representations using pretrained convolutional and transformer-based vision models. Specifically, we consider a ResNet-50 \citep{he2016deep} and a ViT-B/16 \citep{dosovitskiy2020image} architecture, each pretrained on ImageNet. We chose these architectures because they are well-established, state-of-the-art vision models that differ in their inductive biases and representational geometry. RestNet-50 emphasizes local, convolutional feature composition, while ViT-B/16 is transformer-based and relies on global self-attention, allowing us to examine whether different representational assumptions lead to different predictions about exemplar selection. All feature vectors are normalized to have unit norm so that inner products correspond to cosine similarity. Let $f_i$ denote the normalized feature vector associated to the $i$-th image $x_i$. We define the cosine similarity between $x_i$ and $x_j$ as $\mathrm{sim}(i,j) = \langle f_i, f_j \rangle = f_i^T f_j$ along with their cosine distance $\mathrm{dist}(i,j) = 1 - \mathrm{sim}(i,j).$

Leveraging these representations, we will choose a subset of exemplars with quota $M$ that optimizes a specified objective. Each objective aims to mathematically formalize different strategies for choosing exemplars. 


\paragraph{Prototypicality} To identify prototypical exemplars, we rank individual datapoints by their average similarity to the dataset. Given a quota of $M$ exemplars, we then select the top $M$ members according to this score. This criterion identifies stimuli that are considered the most typical based on the neural network's feature representation (e.g., for mammals, dogs, cats, and cows would score higher than whales or bats).

\paragraph{Representativity} To identify exemplar sets that are jointly representative of the dataset, we use a facility location criterion \citep{mirchandani1990discrete, wei2015submodularity} to find a subset $S \subset \{1,\dots,N\}$ of size $|S|=M$ that is most similar on average across the dataset: \begin{align}
    \max_{|S|=M} \mathrm{Representativity}(S):=\sum_{i=1}^N \max_{j \in S}\mathrm{sim}(i,j). \label{eq:similarity}
\end{align} This objective encourages selection of exemplars such that every data point is close to at least one selected exemplar, which implicitly discourages redundancy within the selected exemplars (e.g., it might favor selecting otters and elephants as examples of ``mammal," rather than the more prototypical dogs and cats). Notice that, unlike the prototypicality criterion, this criterion evaluates representativeness at the level of the subset rather than the individual exemplar.


\paragraph{Diversity} To measure diversity independently of representativeness, we select exemplars via \begin{align}
    \max_{|S|=M} \mathrm{Diversity}(S):=\sum_{i < j,i,j \in S} \mathrm{dist}(i,j). \label{eq:diversity}
\end{align} This criterion favors subsets whose members are mutually far apart in the network’s representation space, reflecting the network’s notion of perceptual or semantic diversity. This generally results in better coverage of the category extremes (e.g., selecting whales and bats as ``mammals").

\vspace{-3mm}
\paragraph{Combining objectives} Using the above criteria, one can create hybrid objectives that also combine various attributes. For example, to study trade-offs between representativeness and diversity, we consider a combined objective 
\begin{align}
   \max_{|S|=M} \mathrm{Representativity}(S) + \mathrm{Diversity}(S). \label{eq:combined}
\end{align} This objective aims to balance selecting exemplars that jointly represent the dataset and selecting exemplars that are maximally distinct from one another, weighing each equally.

Together, these selection criteria allow us to disentangle multiple notions of exemplar quality in neural network representations: individual typicality (prototypical exemplars), joint representativeness, and mutual diversity. By applying each criterion to features extracted from different architectures, we can directly compare how neural networks differ in what they consider representative, diverse, or extreme within the same stimulus set.

\vspace{-3mm}

\paragraph{Dependent measures}

We assess the qualities of exemplars selected by both humans and machines through several measures. The first is a \textit{prototypicality score}, which computes the average absolute distance of the exemplars chosen to the prototypical member, which is given by the midpoint. Recall that each member of a category is given a scale value. Given the midpoint's scale value $m_s$, we compute the prototypicality score of a subset of exemplars with scale values $s_1,\dots,s_M$ by \begin{align*}
    \mathrm{prototypicality\ score} = \frac{1}{M}\sum_{i=1}^M \frac{|s_i - m_s|}{m_s}.
\end{align*} We additionally consider a \textit{diversity score} by computing the maximum pairwise distance between exemplars with scale values $s_1,\dots,s_M$ via \begin{align*}
    \mathrm{diversity\ score} = \begin{cases}
        \max_{i \neq j} \frac{|s_i - s_j|}{90} &\ \text{if category is dax or vep,}\\
        \max_{i \neq j} \frac{|s_i - s_j|}{100} &\ \text{if category is bem.} 
    \end{cases}
\end{align*} Note that we normalize both scores (dividing by the category's maximum scale value) so that, regardless of category, they are on the same scale and lie between 0 and 1.

\subsection{Human Participants}

\subsubsection{Participants}
Twenty-four adult participants were recruited from a university participant pool for course credit. 

\subsubsection{Design}
The study was administered in Qualtrics and consisted of three blocks. Each block contained three trials, each featuring a different novel category (see Figure \ref{fig:similarity-all-categories}) and exemplar quota. Thus, within each block, participants viewed all three categories, each assigned a different exemplar quota (1, 2, or 3). Across blocks, each category appeared once in each quota condition. The order of the questions within each block and the presentation of the blocks were randomized.

\subsubsection{Procedure}
Participants were given a cover story in which they were asked to imagine that they were astrobiologists studying aliens from a distant planet. They were told that some aliens had already been reliably identified as members of a given species, and that their task was to teach new trainees how to correctly identify each species.

Participants then completed a series of trials corresponding to the different exemplar-quota conditions. In each trial, they were told, ``These are all [species name, e.g., daxes]. You can choose only [exemplar quota, e.g., one] to teach someone about [daxes]. Which [one] would you pick?'' Participants then selected the exemplar(s) they would use for teaching.

\subsection{Human Results}
We first report results for human participants, before the modeling comparisons. For each participant, we computed a prototypicality score for all exemplar quotas, and the diversity score for 2- and 3-exemplar quotas. We conducted planned analyses to examine whether (1) participants’ prototypicality scores decreased as the number of exemplars available to select increased, (2) participants’ diversity scores exceeded chance, and (3) diversity differed between the 2- and 3-exemplar conditions. We report these analyses in turn. 

\subsubsection{Prototypicality Score}
We first computed chance performance by taking the expected average distance from each stimulus to the category midpoint under uniform random selection with no sensitivity to category structure. This yielded a chance prototypicality score of 23.33 for bems, and 21.11 for veps and daxes. We then normalized the chance prototypicality score by dividing it by the maximum scale value. 

We fit a linear mixed-effects model on chance-centered and normalized prototypicality scores with fixed effects for the exemplar quota condition (treated as a factor with three levels) and random intercepts and slopes for quota by participant. In this measure, higher values indicate greater deviation from the midpoint and thus less prototypical selections. We found a significant effect of exemplar condition (ref = one-exemplar condition), $b = .167$, $SE = .031$, $p < .001$; $b = .163$, $SE = .035$, $p < .001$ for the two- and three-exemplar conditions, respectively. 

We then constructed an intercept-only model for each condition. In the one-exemplar condition, participants' selections did not differ reliably from chance (intercept = -.053, $SE = .036$, $p = .15$). Although the most frequently selected exemplar was the category midpoint (26.4\% of trials), the next most frequent selections were the two end points at 0 and 100 (12.5\% of trials each). This suggests heterogeneity in participants' sample selection strategies: while many favored a central exemplar, a subset of participants preferred boundary cases, exhibiting a diversity-oriented profile. In contrast, prototypicality scores were significantly higher than chance when the participants were allowed to select two ($b = .114$, $SE = 0.02$, $p < .001$) or three exemplars ($b = .109$, $SE = .007$, $p < .001$). This indicates that given the opportunity to select multiple exemplars, adults chose exemplars significantly further from the midpoint than when they were allowed to choose only one exemplar. This preference for selecting exemplars close to the midpoint when given only one exemplar, but exemplars further from the midpoint when given 2 or 3, suggests adults flexibly adapt their sampling strategy across contexts.

\begin{figure}[t]
  \centering
  \includegraphics[width=0.7\columnwidth]{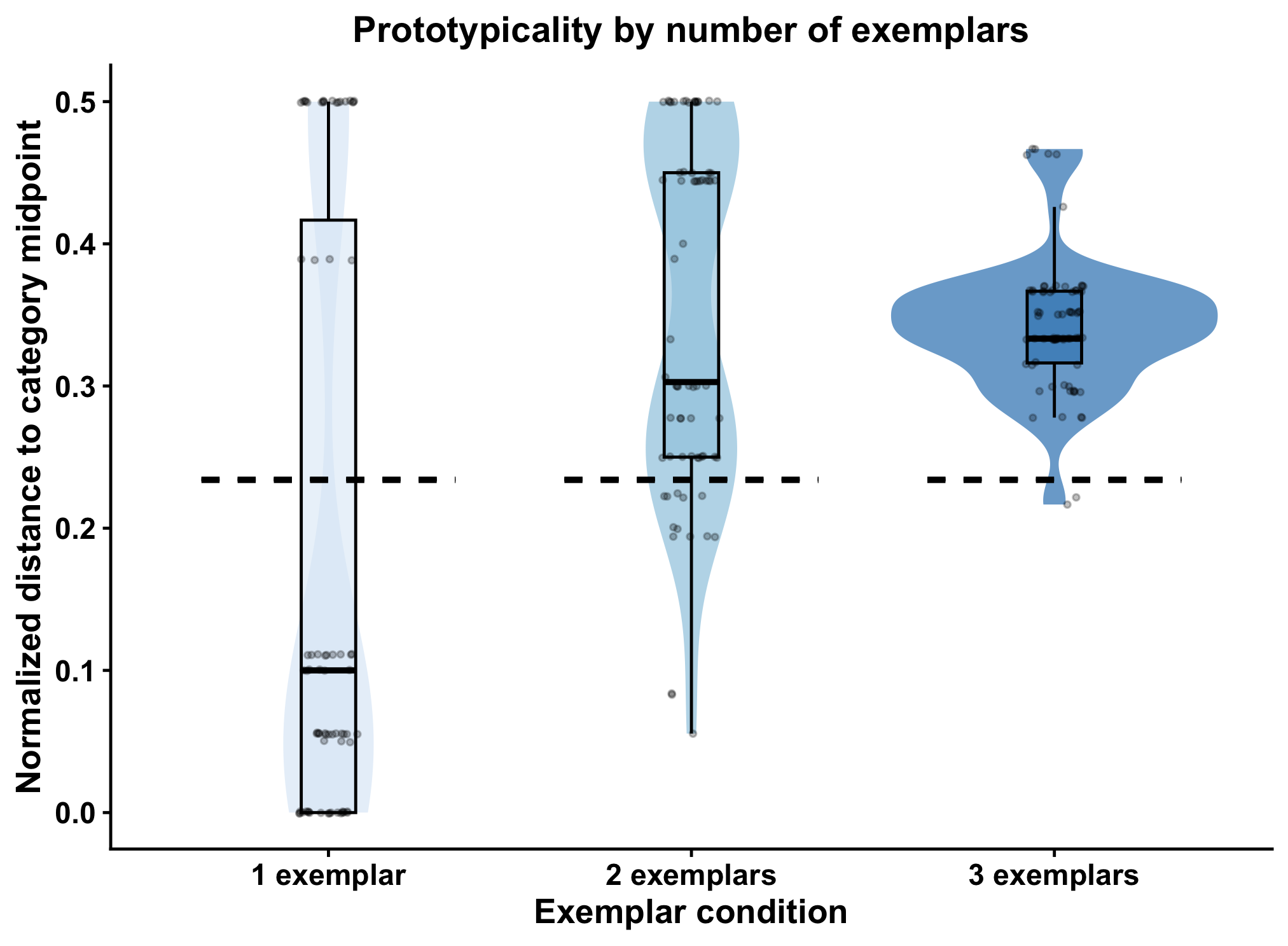}
  \caption{Prototypicality of selected exemplars by condition. Dotted horizontal line indicates chance performance. As the quota for the number of exemplars increases, the prototypicality score tends to increase, meaning that the chosen exemplars become less typical and examples at the endpoints are chosen more frequently.}
  \label{fig:overview}
\end{figure}

\vspace{-5mm}

\paragraph{Diversity Score}
We estimated chance-level diversity in each category and exemplar quota by computing and normalizing the average expected diversity for all possible combinations of exemplars. This yielded an average maximum diversity of .38 for 2-exemplar quotas and .57 for 3-exemplar quotas (as expected, more selections led to greater average diversity). To facilitate comparisons against chance, we subtracted the normalized chance score from the observed score, and ran intercept-only linear mixed-effects models for each quota condition with random intercepts for participants. 

In the two-exemplar condition, the mean chance-centered diversity was significantly greater than zero ($b = .225$, $SE = .05$, $p = .0001$), suggesting that participants' sample selections were on average 22.5\% farther apart on the 0-100\% scale than expected by chance. In the three-exemplar condition, the mean diversity was even greater ($b = .338$, $SE = .023$, $p < .0001$), with participants' selections on average 33.8 percentage points of the category range more diverse than chance level. This is especially notable as the chance-based diversity score was higher in the 3-exemplar condition, yet participants still outperformed it.

We then constructed a linear mixed-effects model predicting chance-centered diversity scores with exemplar quota (2 vs. 3) as a fixed effect and random intercepts and slopes for quota by participant. We observed a significant effect of condition, where the diversity score in the 3-exemplar condition was significantly higher than in the two-exemplar condition, $b = .112$, $SE = .045$, $p = .019$. Thus, when allowed to select three exemplars, instead of two, adults chose even more distinct exemplars.

\subsection{Modeling Results}

We now discuss the modeling results using the neural network features. First, we showcase results from the exemplar selection for each architecture and choice criterion. In particular, Table \ref{tab:model-choices-norm-by-quota} reports the normalized selections of exemplars by architecture, separated by choice criterion and shown alongside the most popular human selection. Across criteria and categories, the models exhibit systematic differences in which exemplars are selected. 

\begin{table*}[t]
\centering
\scriptsize
\setlength{\tabcolsep}{1pt}
\renewcommand{\arraystretch}{0.8}
\begin{tabular}{c|cccc|cccc|c}
\toprule
& \multicolumn{4}{c}{ResNet} & \multicolumn{4}{c}{ViT} & Human \\
Quota & Prototypicality & Representativity & Diversity & Combination & Prototypicality & Representativity & Diversity & Combination & \\
\midrule
1 & 0.5 & 0.5 & 0 & 0.5 & 0.38 & 0.38 & 0 & 0.38 & 0.5 \\
2 & 0.5, 0.5 & 0.38, 0.88 & 0, 1 & 0, 1 & 0.38, 0.5 & 0.11, 0.38 & 0, 1 & 0, 0.4 & 0, 1\\
3 &  0.44, 0.5, 0.55 & 0, 0.5, 0.88 & 0, 0.11, 1 & 0, 0.5, 1 & 0.38, 0.4, 0.5 & 0, 0.5, 0.88 & 0, 0.44, 1 & 0, 0.5, 1 & 0, 0.5, 1\\
\bottomrule
\end{tabular}
\caption{Normalized model exemplar locations by quota. For each quota $M$ and selection criterion, we report the model’s mean normalized exemplar locations (averaged across categories after normalizing scales to $[0,1]$). In the Human column, we report the most popular choice for each quota, also normalized and averaged across categories.}
\label{tab:model-choices-norm-by-quota}
\end{table*}

\begin{figure}[t]
  \centering
  \includegraphics[width=0.7\columnwidth]{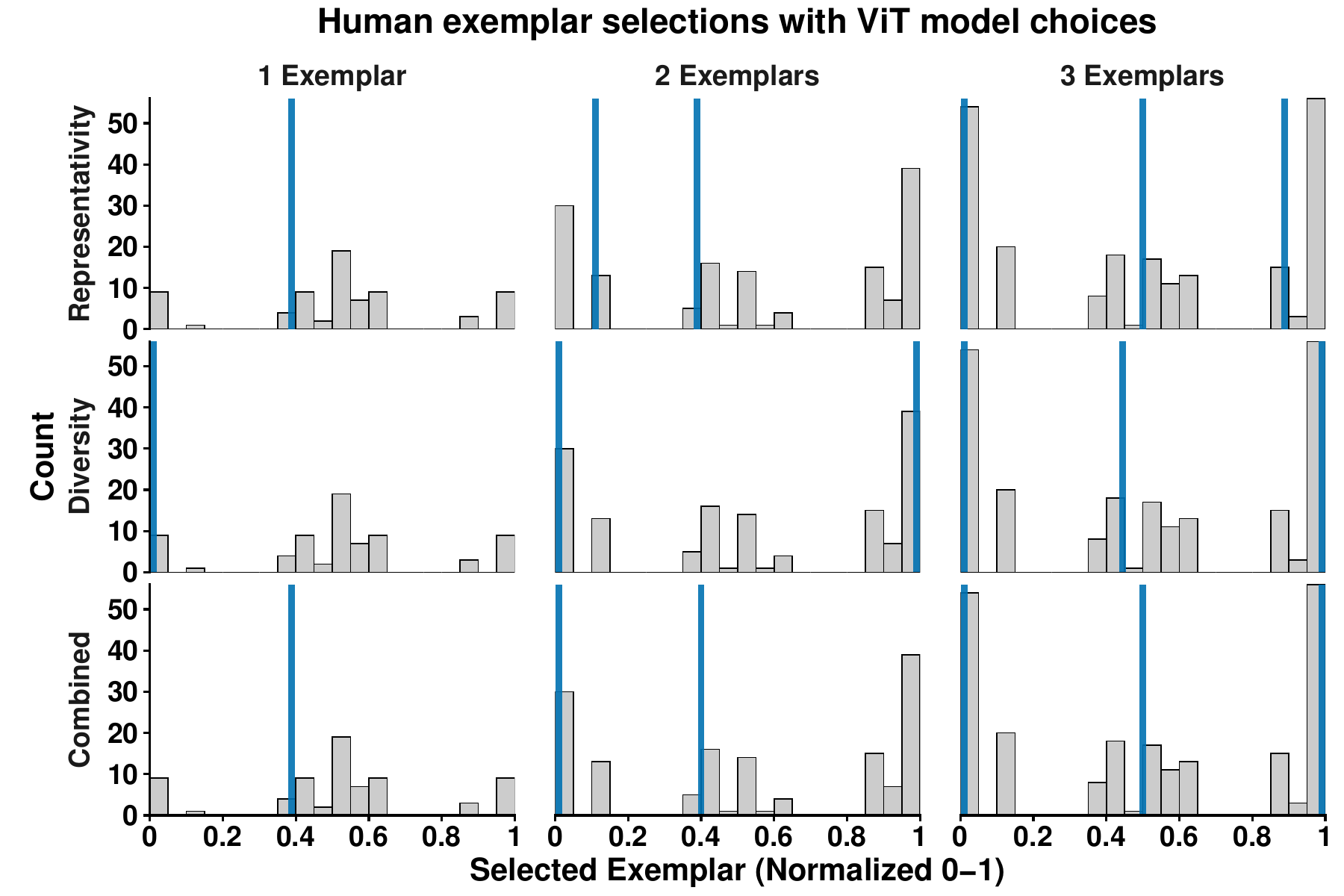}
  \caption{ViT exemplar selection by choice criterion (blue lines) overlaid on human selections.}
  \label{fig:vit-human}
\end{figure}

\subsubsection{Comparison with human data} Next, we consider whether the choices made by the neural network capture the choices made by human participants. To measure accuracy, we compute prototypicality and diversity scores for each model–criterion pair, and then calculate the mean absolute error between the model’s score and each participant’s score in that condition, averaging this error across participants. We conduct this analysis when 1) aggregating over the 1-, 2-, and 3-exemplar conditions and 2) via a more fine-grained analysis comparing model performance in each condition.

\vspace{-3mm}

\paragraph{Aggregating across exemplar conditions}

\begin{figure}[t]
  \centering
  \includegraphics[width=0.7\columnwidth]{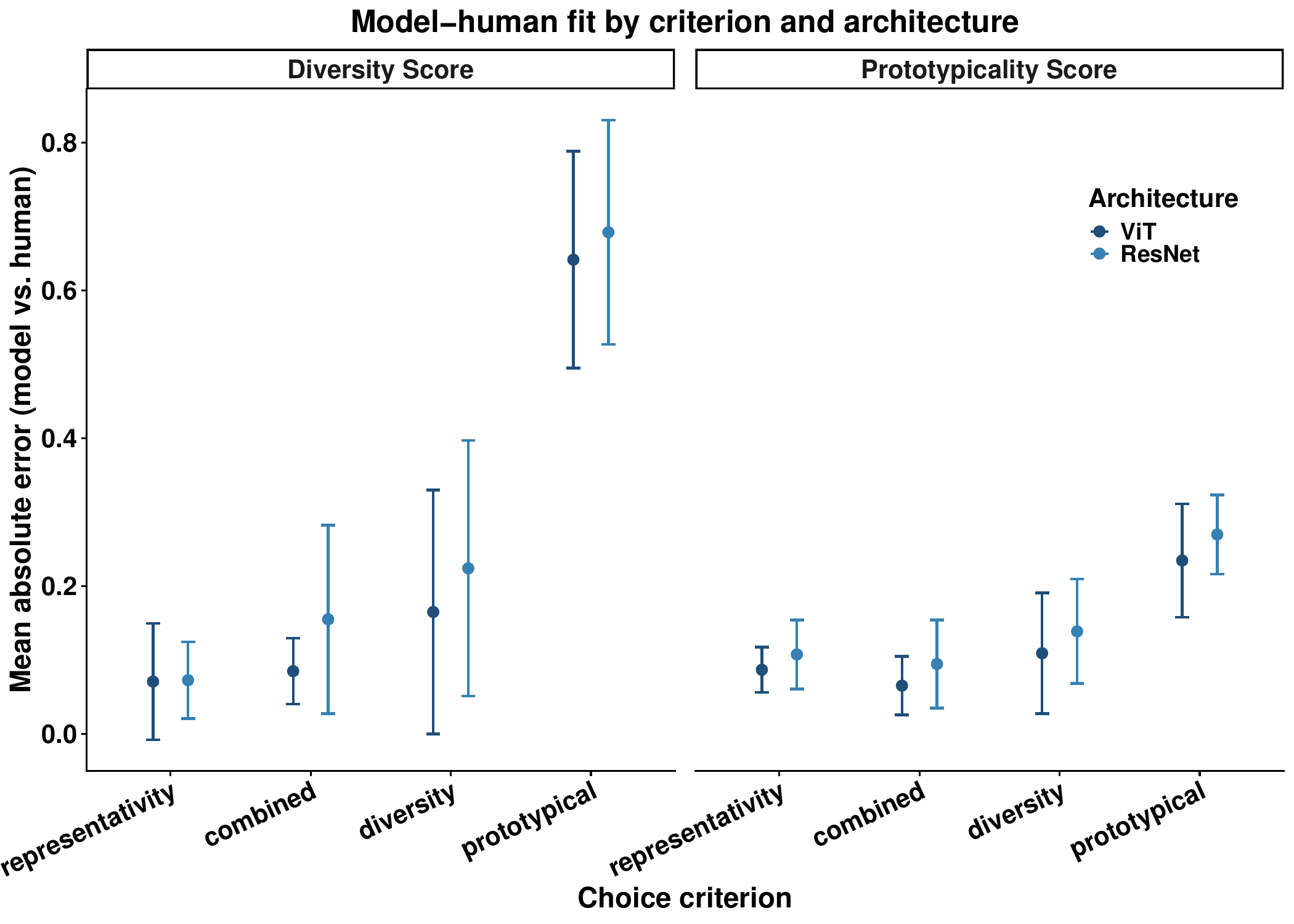}
  \caption{The mean absolute error of the prototypicality score and diversity score between the neural network model and human participants, collapsed across all participants and exemplar quota condition. For the diversity score, the representativity criterion predicts human behavior with an error of $0.071$ and $0.072$ for ViT and ResNet, respectively. For the prototypicality score, the combined criterion yields an error of $0.066$ and $0.095$ for ViT and ResNet, respectively.}
  \label{fig:div-pro-errors}
\end{figure}

We report the mean absolute error of the prototypicality score and diversity score between the neural network predictions and human participants aggregated across exemplar conditions in Figure \ref{fig:div-pro-errors}. In particular, we see that both the ResNet and ViT achieve the lowest mean diversity error (0.072 and 0.071, respectively) and low prototypicality error (0.087 and 0.108, respectively) under the representativity criterion (Eq. \eqref{eq:similarity}). This suggests that selecting exemplars to jointly represent the dataset aligns well with human judgments along both dimensions. Additionally, for ViT, a combined representativeness-diversity criterion (Eq. \eqref{eq:combined}) yields the lowest prototypicality error (0.066), improving over simply representativeness (0.087) or diversity (0.109) individually. Note also that for both models, the diversity criterion (Eq. \eqref{eq:diversity}) is outperformed by representativeness and combined when comparing diversity scores, which seems to indicate that maximizing mutual dissimilarity among exemplars alone does not fully capture human diversity judgments well. Finally, ViT consistently outperforms the ResNet, suggesting its learned representation geometry more closely reflects human representativity judgments in this domain.

\vspace{-5mm}

\paragraph{Fine-grained condition analysis}

\begin{table*}[t]
\centering
\scriptsize
\setlength{\tabcolsep}{2.5pt}
\renewcommand{\arraystretch}{0.9}
\begin{tabular}{c|cccc|cccc}
\toprule
& \multicolumn{4}{c}{ResNet} & \multicolumn{4}{c}{ViT} \\
Quota
& Prototypicality & Representativity & Diversity & Combination
& Prototypicality & Representativity & Diversity & Combination \\
\midrule

\multicolumn{9}{l}{\textbf{Prototypicality score}} \\
\midrule
1
& \underline{0.181} & 0.181 & 0.199 & \underline{0.181}
& \textbf{0.109} & \textbf{0.109} & 0.199 & \textbf{0.109} \\
2
& 0.321 & \underline{0.090} & 0.135 & 0.097
& 0.312 & 0.107 & 0.115 & \textbf{0.070} \\
3
& 0.308 & 0.053 & 0.083 & \textbf{0.007}
& 0.284 & 0.046 & \underline{0.015} & 0.018 \\

\midrule
\multicolumn{9}{l}{\textbf{Diversity score}} \\
\midrule
2
& 0.557 & \underline{0.094} & 0.356 & 0.189
& 0.538 & 0.127 & 0.245 & \textbf{0.055} \\
3
& 0.805 & \underline{0.057} & 0.087 & 0.087
& 0.750 & \textbf{0.020} & 0.062 & 0.062 \\

\midrule
\multicolumn{9}{l}{\textbf{Average Prototypicality + Diversity Score}} \\
\midrule
2
& 0.439 & \underline{0.092} & 0.246 & 0.143
& 0.425 & 0.117 & 0.180 & \textbf{0.063} \\
3
& 0.557 & 0.055 & 0.085 & 0.047
& 0.517 & \textbf{0.033} & \underline{0.039} & 0.040 \\

\bottomrule
\end{tabular}
\caption{
\textbf{Model--human alignment across exemplar quotas and selection criteria.}
We report mean absolute error (MAE) for prototypicality (top), diversity (middle), and their average (bottom).
Bold values indicate the lowest error within each row; underlined values indicate the second lowest.
}
\label{tab:model-errors-combined}
\end{table*}

Table~\ref{tab:model-errors-combined} provides a fine-grained comparison of model-human alignment across exemplar quotas and evaluation metrics. Across conditions, selection strategies based on representativity and a combination of representativity and diversity consistently outperform strategies that optimize prototypicality or diversity alone, particularly in the 2- and 3- exemplar regimes where human diversity behavior is most informative. In particular, the representativity criterion achieves low diversity error while maintaining relatively low prototypicality error, indicating that selecting exemplars to jointly summarize the dataset better reflects human selection strategies. The combined criterion further improves alignment in several cases, especially for ViT, which attains the lowest or second-lowest error across both metrics and quotas. When considering the average of prototypicality and diversity error as a holistic measure of model–human alignment, ViT consistently outperforms ResNet under representativity-based and combined criteria.

To make more direct comparisons with human choices, Figure~\ref{fig:vit-human} provides a qualitative comparison between human exemplar selections and the exemplars chosen by ViT under different choice criteria and exemplar quotas. The diversity criterion tends to focus on extreme endpoints of the morph continuum, while the representativity and combined criterion incorporate choices near the midpoint along with choosing less extreme endpoints (see, e.g., the representativity choices in the 2- and 3- exemplar conditions).

\section{Discussion}

The present work examined how humans select exemplars to teach others and whether principled selection objectives from machine learning can serve as useful computational models of this behavior. Across conditions, our results suggest that human teaching behavior is best characterized not by a single objective, but by a structured balance between representativeness and diversity that shifts with resource constraints.

Selection strategies based on joint representativity (selecting exemplars that collectively summarize the dataset) provide a closer match to human behavior on both prototypicality and diversity measures than strategies that prioritize either individual prototypicality or mutual dissimilarity alone. This suggests adults are structuring selections to jointly represent the distribution. Hybrid objectives that combine representativeness and diversity further improve alignment with human judgments. This finding supports the idea that human teachers may implicitly trade off multiple pedagogical goals, rather than optimizing a single criterion. We also found that the hybrid objective criterion closely parallels participants' modal selections (Table~\ref{tab:model-choices-norm-by-quota}). While a representativity criterion rarely selects exemplars at the extremes (preferring not-quite-extreme exemplars), human learners often select extreme exemplars in both 2- and 3-exemplar conditions. This pattern is captured well by the objective combining representativity and diversity.

Another notable result is the consistent advantage of transformer-based representations over convolutional ones in predicting human behavior. Across criteria, ViT-based models achieved lower error relative to human judgments than ResNet-based models. While both architectures are high-performing vision models, they differ substantially in their inductive biases: convolutional networks emphasize local feature composition, whereas transformers rely on global self-attention. Our findings suggest that the representational geometry induced by global attention may better capture the similarity relations humans use to evaluate exemplars on a continuum. An important direction for future work is to examine a broader range of architectures and inductive biases to better understand which representational properties are critical for capturing human exemplar selection.

Beyond the specific results, this work highlights the value of exact, interpretable subset selection objectives as cognitive models, with an emphasis on interpretability and theoretical transparency. However, the neural network representations we use remain complex and difficult to interpret directly. It would be of interest to develop smaller or more interpretable representational models that can capture human exemplar selection behavior while offering greater explanatory insight.

A second promising direction concerns individual differences in exemplar selection. As noted above, there is some heterogeneity across learners (especially in 1-exemplar contexts); subsequent work might model individual learners and test how strategies shift based on learning contexts. Future work might also consider how such strategies emerge over development, building on findings that younger children prioritize category typicality over diversity \citep{rhodes2008sample}.

Finally, our work suggests potential avenues for connections to recent advances in synthetic dataset generation in machine learning. Work in dataset distillation has shown that for machine learning models, generating small, synthetic datasets outperform using subsets of real examples for downstream learning \citep{tukan2023dataset, sajedi2023datadam, sun2024diversity}. An open question is whether analogous synthetic exemplars optimized to convey category structure while controlling for realism could serve as effective teaching examples for humans. Exploring whether synthetic exemplars can improve human learning, or how humans respond to such examples, could open a new avenue for studying the computational principles underlying teaching and information transmission.

\bibliographystyle{plain}

\bibliography{main}

@article{shafto2014rational,
  title={A rational account of pedagogical reasoning: Teaching by, and learning from, examples},
  author={Shafto, Patrick and Goodman, Noah D and Griffiths, Thomas L},
  journal={Cognitive psychology},
  volume={71},
  pages={55--89},
  year={2014},
  publisher={Elsevier}
}

@article{rhodes2010children,
  title={Children’s attention to sample composition in learning, teaching and discovery},
  author={Rhodes, Marjorie and Gelman, Susan A and Brickman, Daniel},
  journal={Developmental Science},
  volume={13},
  number={3},
  pages={421--429},
  year={2010},
  publisher={Wiley Online Library}
}

@article{avrahami1997teaching,
  title={Teaching by examples: Implications for the process of category acquisition},
  author={Avrahami, Judith and Kareev, Yaakov and Bogot, Yonatan and Caspi, Ruth and Dunaevsky, Salomka and Lerner, Sharon},
  journal={The Quarterly Journal of Experimental Psychology Section A},
  volume={50},
  number={3},
  pages={586--606},
  year={1997},
  publisher={SAGE Publications Sage UK: London, England}
}

@article{csibra2011natural,
  title={Natural pedagogy as evolutionary adaptation},
  author={Csibra, Gergely and Gergely, Gy{\"o}rgy},
  journal={Philosophical Transactions of the Royal Society B: Biological Sciences},
  volume={366},
  number={1567},
  pages={1149--1157},
  year={2011},
  publisher={The Royal Society}
}

@article{latourrette2022sparse,
  title={Sparse labels, no problems: Infant categorization under challenging conditions},
  author={LaTourrette, Alexander and Waxman, Sandra R},
  journal={Child development},
  volume={93},
  number={6},
  pages={1903--1911},
  year={2022},
  publisher={Wiley Online Library}
}

@inproceedings{he2016deep,
  title={Deep residual learning for image recognition},
  author={He, Kaiming and Zhang, Xiangyu and Ren, Shaoqing and Sun, Jian},
  booktitle={Proceedings of the IEEE conference on computer vision and pattern recognition},
  pages={770--778},
  year={2016}
}

@article{lei2023comprehensive,
  title={A comprehensive survey of dataset distillation},
  author={Lei, Shiye and Tao, Dacheng},
  journal={IEEE Transactions on Pattern Analysis and Machine Intelligence},
  volume={46},
  number={1},
  pages={17--32},
  year={2023},
  publisher={IEEE}
}

@article{tsang2005core,
  title={Core vector machines: Fast SVM training on very large data sets.},
  author={Tsang, Ivor W and Kwok, James T and Cristianini, Nello and others},
  journal={Journal of Machine Learning Research},
  volume={6},
  number={4},
  year={2005}
}

@article{har2007smaller,
  title={Smaller coresets for k-median and k-means clustering},
  author={Har-Peled, Sariel and Kushal, Akash},
  journal={Discrete and Computational Geometry},
  volume={37},
  number={1},
  pages={3--19},
  year={2007},
  publisher={Springer New York}
}

@article{olvera2010review,
  title={A review of instance selection methods},
  author={Olvera-L{\'o}pez, J Arturo and Carrasco-Ochoa, J Ariel and Mart{\'\i}nez-Trinidad, J Francisco and Kittler, Josef},
  journal={Artificial Intelligence Review},
  volume={34},
  number={2},
  pages={133--143},
  year={2010},
  publisher={Springer}
}

@article{wang2018dataset,
  title={Dataset distillation},
  author={Wang, Tongzhou and Zhu, Jun-Yan and Torralba, Antonio and Efros, Alexei A},
  journal={arXiv preprint arXiv:1811.10959},
  year={2018}
}

@inproceedings{sucholutsky2021secdd,
  title={Secdd: Efficient and secure method for remotely training neural networks (student abstract)},
  author={Sucholutsky, Ilia and Schonlau, Matthias},
  booktitle={Proceedings of the AAAI Conference on Artificial Intelligence},
  volume={35},
  number={18},
  pages={15897--15898},
  year={2021}
}

@article{zhao2020dataset,
  title={Dataset condensation with gradient matching},
  author={Zhao, Bo and Mopuri, Konda Reddy and Bilen, Hakan},
  journal={arXiv preprint arXiv:2006.05929},
  year={2020}
}

@inproceedings{dong2022privacy,
  title={Privacy for free: How does dataset condensation help privacy?},
  author={Dong, Tian and Zhao, Bo and Lyu, Lingjuan},
  booktitle={International Conference on Machine Learning},
  pages={5378--5396},
  year={2022},
  organization={PMLR}
}

@article{havy2016naming,
  title={Naming influences 9-month-olds’ identification of discrete categories along a perceptual continuum},
  author={Havy, M{\'e}lanie and Waxman, Sandra R},
  journal={Cognition},
  volume={156},
  pages={41--51},
  year={2016},
  publisher={Elsevier}
}

@article{dosovitskiy2020image,
  title={An image is worth 16x16 words: Transformers for image recognition at scale},
  author={Dosovitskiy, Alexey},
  journal={arXiv preprint arXiv:2010.11929},
  year={2020}
}

@book{mirchandani1990discrete,
  title={Discrete location theory},
  author={Mirchandani, Pitu B and Francis, Richard L},
  year={1990}
}

@inproceedings{wei2015submodularity,
  title={Submodularity in data subset selection and active learning},
  author={Wei, Kai and Iyer, Rishabh and Bilmes, Jeff},
  booktitle={International conference on machine learning},
  pages={1954--1963},
  year={2015},
  organization={PMLR}
}

@inproceedings{sener2018active,
  title={Active Learning for Convolutional Neural Networks: A Core-Set Approach},
  author={Sener, Ozan and Savarese, Silvio},
  booktitle={International Conference on Learning Representations},
  year={2018}
}

@article{li2023bal,
  title={Bal: Balancing diversity and novelty for active learning},
  author={Li, Jingyao and Chen, Pengguang and Yu, Shaozuo and Liu, Shu and Jia, Jiaya},
  journal={IEEE Transactions on Pattern Analysis and Machine Intelligence},
  volume={46},
  number={5},
  pages={3653--3664},
  year={2023},
  publisher={IEEE}
}

@article{rhodes2008sample,
  title={Sample diversity and premise typicality in inductive reasoning: Evidence for developmental change},
  author={Rhodes, Marjorie and Brickman, Daniel and Gelman, Susan A},
  journal={Cognition},
  volume={108},
  number={2},
  pages={543--556},
  year={2008},
  publisher={Elsevier}
}

@article{tukan2023dataset,
  title={Dataset distillation meets provable subset selection},
  author={Tukan, Murad and Maalouf, Alaa and Osadchy, Margarita},
  journal={arXiv preprint arXiv:2307.08086},
  year={2023}
}

@inproceedings{sajedi2023datadam,
  title={Datadam: Efficient dataset distillation with attention matching},
  author={Sajedi, Ahmad and Khaki, Samir and Amjadian, Ehsan and Liu, Lucy Z and Lawryshyn, Yuri A and Plataniotis, Konstantinos N},
  booktitle={Proceedings of the IEEE/CVF International Conference on Computer Vision},
  pages={17097--17107},
  year={2023}
}

@inproceedings{sun2024diversity,
  title={On the diversity and realism of distilled dataset: An efficient dataset distillation paradigm},
  author={Sun, Peng and Shi, Bei and Yu, Daiwei and Lin, Tao},
  booktitle={Proceedings of the IEEE/CVF Conference on Computer Vision and Pattern Recognition},
  pages={9390--9399},
  year={2024}
}

\end{document}